# CytoCLIP: Learning Cytoarchitectural Characteristics in Developing Human Brain Using Contrastive Language Image Pre-Training

Pralaypati Ta[a,b], Sriram Venkatesaperumal[a], Keerthi Ram[a], and Mohanasankar Sivaprakasam[a,b]

[a] Sudha Gopalakrishnan Brain Centre, Indian Institute of Technology, Madras, Chennai, India.
[b] Department of Electrical Engineering, Indian Institute of Technology, Madras, Chennai, India.

Corresponding author: Pralaypati Ta (e-mail: pralaypati@htic.iitm.ac.in)

***Abstract*— The functions of different regions of the human brain are closely linked to their distinct cytoarchitecture, which is defined by the spatial arrangement and morphology of the cells. Identifying brain regions by their cytoarchitecture enables various scientific analyses of the brain. However, delineating these areas manually in brain histological sections is time-consuming and requires specialized knowledge. An automated approach is necessary to minimize the effort needed from human experts. To address this, we propose CytoCLIP, a suite of vision-language models derived from pre-trained Contrastive Language-Image Pre-Training (CLIP) frameworks to learn joint visual-text representations of brain cytoarchitecture. CytoCLIP comprises two model variants: one is trained using low-resolution whole-region images to understand the overall cytoarchitectural pattern of an area, and the other is trained on high-resolution image tiles for detailed cellular-level representation. The training dataset is created from NISSL-stained histological sections of developing fetal brains of different gestational weeks. It includes 86 distinct regions for low-resolution images and 379 brain regions for high-resolution tiles. We evaluate the model's understanding of the cytoarchitecture and generalization ability using region classification and cross-modal retrieval tasks. Multiple experiments are performed under various data setups, including data from samples of different ages and sectioning planes. Experimental results demonstrate that CytoCLIP outperforms existing methods. It achieves a weighted F1 score of 0.87 for whole-region classification and 0.91 for high-resolution image tile classification.**

## I. INTRODUCTION

THE human brain is a complex system with many interconnected structures. Examining its structural and functional organization is essential for advancing the treatment of neurological disorders, understanding human behaviour, and cognition. The brain consists of various regions, each showing distinct cytoarchitectural characteristics determined by the spatial organization of cells, including their shape, density, size, type, and arrangement. The specific cytoarchitecture of each brain area plays a crucial role in its functionality (Zilles & Amunts, 2010; Weiner et al., 2017). Identifying these distinct characteristics across the different brain regions enables functional, molecular, and connectivity analyses of the brain. NISSL-stained high-resolution images of brain histological sections, such as those in DHARANI (Verma et al., 2025) and the Allen Brain Atlas (Ding et al., 2022), enable the delineation of brain regions based on their cytoarchitecture. However, such a delineation process usually requires the involvement of human experts to interpret the images, making it both costly and time-consuming. An automated approach supplementing the interpretation of human experts at scale is needed (Das et al., 2025; Schiffer, Spitzer, et al., 2021).

Previous research has attempted to automate the segregation of brain areas based on their cytoarchitectural characteristics (Schiffer, Spitzer, et al., 2021; Schiffer, Amunts, et al., 2021; Spitzer et al., 2018). However, these efforts generally involved a limited number of regions, and the accuracy of the output is also low. Recently, vision-language models trained using contrastive learning have demonstrated strong performance in various histopathology image processing tasks, including image classification, captioning, and cross-modal retrieval (Lu et al., 2024; Ciga et al., 2022). Additionally, biomedical domain-specific models based on the Contrastive Language Image Pre-Training (CLIP) (Radford et al., 2021) framework have also demonstrated impressive results in these tasks (Kakkar et al., 2024; Zhang et al., 2025; Eslami et al., 2023; Wang et al., 2022).

In this work, we investigate the use of contrastive learning to learn the cytoarchitectural characteristics of brain regions in histological images of the developing human brain. We develop two varieties of models, called CytoCLIP, from pre-trained base CLIP models that understand the relationship between brain regions and their corresponding cytoarchitecture. The models

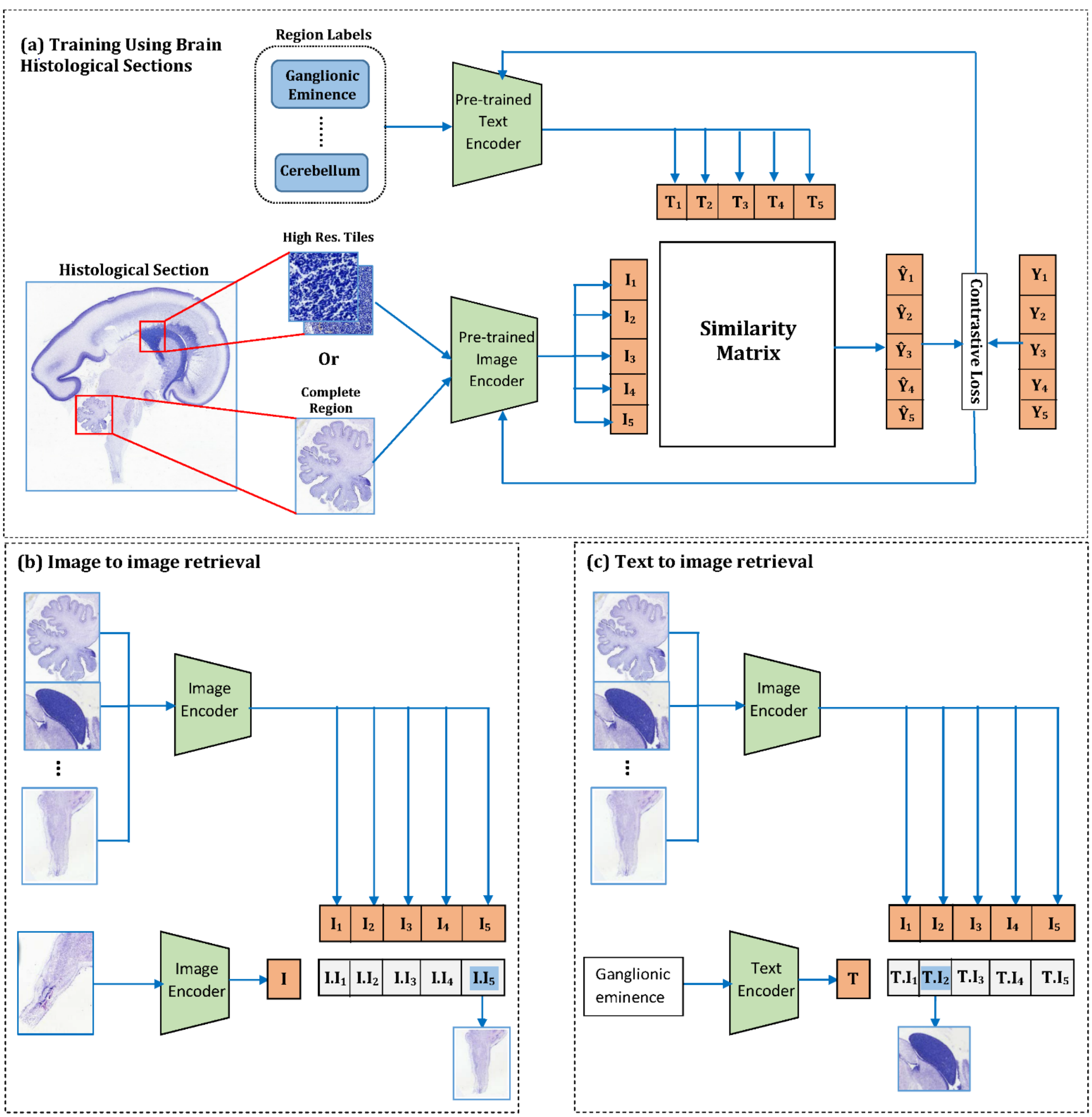

Fig. 1 (a) Training CLIP model using NISSL-stained brain histological sections. (b) Image-To-Image retrieval using CytoCLIP: Retrieval of brain regions using region image as query. (c) Text-To-Image retrieval using CytoCLIP: Retrieval of brain regions using region name as query.

are trained on NISSL-stained histological section images from the DHARANI dataset (Verma et al., 2025). They are aimed at learning cytoarchitectural characteristics for a large number of brain regions. The first type of model is trained using low-resolution images of complete brain regions. These models aim to capture the broader cytoarchitectural characteristics of brain regions from the low-resolution images. The second type of model is trained on very high-resolution image tiles that provide detailed cytoarchitectural patterns to differentiate brain regions. The training method of the CytoCLIP models is shown in Fig. 1(a). We evaluate the performance of the models on region prediction, framed as a multi-class classification problem, as well as cross-modal retrieval tasks. The models show strong performance compared to existing works on both these tasks. Our key contributions are as follows:

- We prepare datasets from NISSL-strained histological sections of brains for low-resolution whole region images and high-resolution tiles with a diverse array of distinct regions. The low-resolution whole region dataset includes 86 distinct regions, whereas the high-resolution region tile dataset contains 379 distinct regions.

- We develop three different models from pre-trained CLIP models using the above dataset.
- We evaluate the performance of the trained models on region classification and cross-modal retrieval tasks across various data setups, such as low and high-resolution histological images, single and multi-region labels. Furthermore, we evaluate the generalization ability of the trained models across the ages of the brain samples and the histological sectioning planes using various cross-data setups.

## II. Related Work

In this work, we adapt the Contrastive Language Image Pre-Training (CLIP) (Radford et al., 2021) model to learn cytoarchitectural features from brain histological section images. We first discuss the related works that employ contrastive learning and CLIP-based models for biomedical and histological image processing tasks. Next, we discuss the works that utilize cellular-level details to classify and segment images of brain histological sections.

### A. Machine Learning for Histological Image Analysis

Deep learning methods have gained significant traction for analyzing histopathological whole-slide images (WSI). They aid in disease classification and detection, segmentation of the region of interest (ROI), retrieval of similar images, and numerous other tasks (Komura & Ishikawa, 2018; Madabhushi & Lee, 2016). Various deep learning architectures and methodologies have been proposed for different histopathology image analysis tasks (Hu et al., 2023; Kosaraju et al., 2022). Recently, Vision Transformers (ViTs) showed promising results in classification, segmentation, and other tasks related to histopathological WSI processing (Hu et al., 2023). It uses the multi-head self-attention mechanism to extract more robust features. UNI (R. J. Chen et al., 2024) trained a large ViT architecture using self-supervised learning on over 100 million WSIs to develop a self-supervised vision encoder for pathology. HIPT (R. J. Chen et al., 2022) learns high-resolution representations of WSIs through knowledge-distillation using the hierarchical structure of WSIs.

### B. Contrastive Learning for Histology Images

Contrastive learning is widely used in the processing of histopathology images (Ciga et al., 2022; Fashi et al., 2022; Rahaman et al., 2025). It extracts task-agnostic & meaningful representations by differentiating between similar and dissimilar image pairs (T. Chen et al., 2020). Ciga et al. (2022) employs the contrastive learning method, as proposed in SimCLR (T. Chen et al., 2020), to learn domain-specific features from histopathology images. It generates similar images by generating augmented views from the same image data. A self-supervised triplet contrastive learning method, along with image masking, is used by Zhao et al. (2023) to classify endometrial histopathological images. Yang et al. (2021) proposes the stain vector perturbation method for data augmentation and uses cross-stain prediction in conjunction with contrastive learning to learn visual representation from histopathology images. The geodesic distance, as a similarity metric, is used with contrastive learning by Tan & Jeong (2023) to generate better feature representations of histopathology WSIs. It results in better classification performance than the cosine distance-based metric.

### C. Contrastive Vision-Language Pretraining

Contrastive Vision-Language Models have dual encoders that embed images and texts into a shared embedding space. As in CLIP (Radford et al., 2021), given a large dataset, the encoders employ contrastive learning during pretraining and learn to generate embeddings with high similarity scores for matching image-text pairs and low similarity scores for non-matching pairs. Several variants of CLIP have been used in medical imaging (Zhao et al., 2025) as well as for histopathology image processing (Lu et al., 2023; Patel et al., 2024). Patel et al. (2024) demonstrates that models trained on domain-specific datasets, such as PLIP (Huang et al., 2023) and BiomedCLIP (Zhang et al., 2025), achieve superior performance in hematology compared to CLIP (Radford et al., 2021), which is trained on a generic dataset. CONCH (Lu et al., 2024) is trained on millions of histopathological images and biomedical text pairs collected from various sources. It achieves superior performance on histopathology image classification, segmentation, multi-modal retrieval, and many other tasks. KL-CVR (Wei et al., 2025) has explored the use of domain-specific medical knowledge, along with vision–language contrastive learning, for biomedical image classification and image-to-image retrieval. This approach shows better performance compared to the original CLIP (Radford et al., 2021) and remains competitive with other state-of-the-art models, even with a smaller dataset. In this work, we train CLIP (Radford et al., 2021) and its variant BiomedCLIP (Zhang et al., 2025) on a dataset created from brain histological sections.

### D. Cytoarchitecture Mapping of Brain Histology Images

Various research studies have investigated the mapping of brain regions based on their cytoarchitectural characteristics (Zilles & Amunts, 2010; Amunts & Zilles, 2015). Given the complexity of the human brain, the cytoarchitectural mapping often requires manual delineation of the brain regions based on neuronal cell distribution, density, and other characteristics (Weiner et al., 2017; Verma et al., 2025; Ding et al., 2022; Rottschy et al., 2007; Ding et al., 2016). However, with the rise of deep learning, several research works have begun to explore cytoarchitectural mapping using deep learning techniques. Spitzer et al. (2017) uses a deep convolutional neural network (CNN) for automatically parcellating cortical areas in high-resolution brain histological sections. It shows that learning of the model is transferable to new brains and also consistent across the sections within the same brain. The work in Schiffer, Spitzer, et al. (2021) employs a CNN-based workflow to automatically classify brain areas on a large scale based on cytoarchitectonic features. It develops a set of separate CNNs, each specialized for automatically mapping the regions from a specific series of histological sections. To avoid the requirement of extensive annotated data for training CNN networks, a self-supervised Siamese network is used in Spitzer et al. (2018) for cytoarchitectonic segmentation of brain regions. Their experiment shows that the use of a self-supervised Siamese network significantly improves the Dice score. Schiffer, Amunts, et al. (2021) employs contrastive learning for the cytoarchitectonic mapping of brain histological sections.

## III. Methods

The dataset used to train the models in this work is generated by processing the images of the brain histological sections available in the DHARANI dataset (SGBC-IITM, 2025). We will first briefly discuss the DHARANI dataset, then outline the methodology for creating the dataset for this work, and finally, the training methodology will be discussed.

### *A. DHARANI dataset*

The DHARANI dataset (SGBC-IITM, 2025) includes the images of brain histological sections created from 5 fetal brain specimens at 14 to 24 gestational weeks (GW). Out of the five fetal brains, four brains were sectioned in the Sagittal plane, whereas the other one was sectioned in the Coronal plane. The histological sections were digitized at 0.5 µm/pixel. The DHARANI dataset contains 466 Nissl-stained, annotated sections, which are annotated with 414 identified brain structures, including brain regions, fiber tracts, ventricles, and other developmental structures. In this work, we generated the dataset from the downsampled version of the annotated histological section images having a resolution of 2 µm/pixel and 16 µm/pixel.

The section images are annotated using a hierarchical nomenclature (SGBC-IITM, 2025). The topmost categories in the nomenclature are *brain*, *spinal cord*, *fiber tracts*, *developmental structures*, and *ventricles*. The sub-regions are organized under these categories at different levels. For example, the *brain* and *spinal cord* categories include all the regions in the gray matter parts and their parents of the human developing brain. For some of the categories, the depth of the corresponding nomenclature subtree is as high as eight.

### *B. Data preparation for low-resolution (16 µm/pixel) whole region images*

We did not consider all regions while compiling the dataset from the low-resolution whole-region images, particularly the areas that spread across the section images. Additionally, the leaf-level regions in these low-resolution images are small and lack substantive information. Therefore, we merged the sub-regions at various intermediate hierarchy levels for specific brain regions and labeled them according to their parent regions.

*Ventricles* and *White matter fibers* are spread across the various parts of a section image. The bounding box that includes these regions spans a larger area of the section image and incorporates numerous neighboring areas. Hence, we did not consider these two regions and their children while preparing the dataset for training with low-resolution images.

In the nomenclature tree, the children of the *Forebrain* are *Telencephalon* and *Diencephalon*. Under the *Forebrain* subtree, we merged regions two levels below the *Telencephalon* and *Diencephalon* with their corresponding parents, except for the subtree under the *Cerebral cortex*. The above merging resulted in the following major regions: *Amygdala*, *Basal nuclei*, *Lateral migratory stream*, *Rostral migratory stream*, *Ganglionic eminence*, *Thalamus*, and *Hypothalamus*. Under the Cerebral cortex, the transient layers of the different developing cortices are mapped to the corresponding parent. The other major regions under the Cerebral cortex subtree, whose children are merged into it, are as follows: *Hippocampus*, *Subiculum*, *Presubiculum*, *Parasubiculum*, and *Postsubiculum*.

For the subtrees under the *Midbrain*, *Hindbrain*, *Fiber tracts*, and *Developmental (transient) structures*, the sub-regions two level below these regions have been merged into their corresponding parents. These merging results in significant areas such as *Pretectum*, *Tectum*, *Tegmentum*, *Pons*, *Medulla*, *Corticofugal tract*, etc. Finally, the children of *Cerebellum*, *Cerebellar nuclei*, *Germinal trigone*, and *Spinal cord* are merged into their corresponding parents.

Certain areas in a histological section can be very thin or small based on the location of the section in the brain. For example, this can occur in regions like the *Internal medullary lamina of the thalamus or Rhombencephalic neuroepithelium*. We discarded such areas as they do not contain any significant information. We calculated the area and the length of the perimeter of the boundary polygon of regions at 16 microns per pixel resolution. We discarded the region if the area is less than a certain threshold. Also, if the area-to-perimeter ratio is less than a certain threshold, the region is considered narrow, and we discard it.

Also, based on the location of the histological section in the brain, a few merged areas can have more than one part and multiple polygons as we merged low-level regions in the nomenclature tree to their parents, e.g., the *Corticofugal tract* and a few other areas has many such parts in a few sections. If these polygons are close to each other, we combine them and create a single image. The shortest distance between each pair of polygons is calculated first. Then, starting with a particular polygon, we performed a proximity-based depth-first search to find the polygons that are close to each other. A threshold distance value is used to determine the proximity of the polygons. If all the parts are close enough, we generate a single image for a region that includes the complete area. In some cases, the parts are not close to each other. We created multiple images for such brain regions and added a part number on the label of each part image.

We cropped a complete brain region from the histological section image using the bounding box derived from its final polygon. Multiple images were generated for a brain region by applying different cropping techniques: 1) **ExactBBox**: cropping based on the exact bounding box, which is generally rectangular, that tightly encloses a region polygon, and 2) **SquareBBox**: cropping based on a square bounding box. The square bounding box is created by expanding the original bounding box to encompass portions of neighboring regions. We also generated an image mask based on the exact boundary of the polygon for the **ExactBBox** types of images. The mask can be applied to get the exact region image without any neighboring region. For the **SquareBBox** type of images, we generated two versions with minimum dimensions of 336 pixels and 224 pixels. This process resulted in three datasets containing 13618 region images of Sagittal sectioning, 5080 images of Coronal sectioning, and 86 distinct region names in each set. The region names are used as the primary label for the images. We also assigned another set of labels, called *multi-region* label, for the **SquareBBox** types of images by including the neighboring regions' names and the primary label where more than 80% of a neighboring region is included in the image.

### C. Data preparation for high-resolution (2 µm/pixel) images

The high-resolution dataset for this work was generated from the histological section images having a resolution of 2 µm/pixel. We generated tiles of size 224 x 224 pixels from a Whole Slide Image (WSI). The boundary of each tile is compared against the boundary of the leaf-level brain structures in the nomenclature tree, and the portion of the tile that overlaps with the region is calculated. If the tile overlaps with a region by more than 40 percent of its area, the region name is considered a potential label for that tile. If there is more than one such label, the region's name with maximum overlapping with the tile is regarded as the label for that tile. This process resulted in 3.34 million tiles from the sagittal sectioned images and 0.96 million from the coronal sectioned images. The tiles are labeled with 379 distinct brain structures.

### D. Training details and Hyperparameters

We trained three base models using the datasets mentioned above: OpenAI's clip-vit-large-patch14-336 and clip-vit-large-patch14 (Radford et al., 2021) models and BiomedCLIP-PubMedBERT_256-vit_base_patch16_224 (Zhang et al., 2025) from Microsoft. The OpenAPI models were trained using the Huggingface transformer library, whereas BiomedCLIP was trained using the open_clip (Ilharco et al., 2025) library. All three models were trained using the whole-region images, and only BiomedCLIP was trained using high-resolution tiles. For OpenAI models, we used the region names as the text prompt, whereas for BiomedCLIP, the prompt template "*this is a photo of <region name>*" was used. The Symmetric Cross-Entropy loss (Radford et al., 2021), mentioned below, was used as the training objective.

$$\mathcal{L} = \frac{1}{2N}\sum_{i=1}^{N}\left(-log\left(\frac{exp\left(\tau \cdot sim(I_i,T_i)\right)}{\sum_{j=1}^{N} exp\left(\tau \cdot sim\left(I_i,T_j\right)\right)}\right)\right) + \frac{1}{2N}\sum_{i=1}^{N}\left(-log\left(\frac{exp\left(\tau \cdot sim(T_i,I_i)\right)}{\sum_{j=1}^{N} exp\left(\tau \cdot sim\left(T_i,I_j\right)\right)}\right)\right) \qquad (1)$$

where $N$ is the number of image-text pairs, $\tau = exp(logit_scale)$ a learnable temperature parameter, $sim\left(I_i,T_j\right)$ is the cosine similarity between the embeddings of image $I_i$ and text $T_j$.

We used the AdamW optimizer (Loshchilov & Hutter, 2019) with a base learning rate of 5e-5 for the training. Each model was trained for 50 epochs. For whole region images, the models were trained using a single NVIDIA A100 80GB GPU, and for high-resolution images, 8 NVIDIA A100 80GB GPUs were used to train the BiomedCLIP model. Using a batch size of 64, it took 20 hours to complete the 50 epochs for training using the high-resolution tiles.

### E. Image transformation

Depending on the model's requirement, the low-resolution complete region images must be converted to 336x336 or 224x224 pixel dimensions. Whenever required, we resized the images using the bicubic sampling. The **ExactBBox** complete region images described above generally do not have a square shape. To avoid any information loss from the cropping applied by the model's default image transformation algorithm, we applied a transformation algorithm to convert the images to 336x336 or 224x224 by applying padding with pixels having zeros color values in all channels. For large brain regions such as the cortex, it resulted in a reduction of the image dimensions by up to a factor of 10. The high-resolution tiles are exactly 224 x 224 dimensions, as expected by the model; no additional transformation was applied. The default transformation of the model was applied to the images.

### *F. Test dataset*

The test datasets were generated following the 80:20 rule. It includes the images from different histological sections to evaluate the model's generalization capability across diverse section images. In the low-resolution whole region dataset, a particular region label occurs only once in a section image. For one such region label, we randomly selected 20% of all the sections containing the particular brain region. The images from all these sections are used as test data for that region, allowing us to assess whether the model can recognize the region from other sections not included in the training data.

In high-resolution WSI tiles, a region spans multiple tiles within a single section image. For each region label, we randomly selected 20% of all tile images generated across all sections corresponding to that label. These selected tile images are used as test data for that label. Therefore, the test data for a region consists of tiles from other parts of the same section, along with the tiles from different sections. It helps to analyze the model's ability to generalize to other areas of the same region.

TABLE 1
DETAILS OF THE DATASET USED FOR TRAINING, VALIDATION, AND TEST.

| Image Type | Dataset # | Image Size | Cropping Method | Labels Count | Label Type | Sectioning Plane | Brain Specimens |
|---|---|---|---|---|---|---|---|
| Complete Region | 1 | 224 | SquareBBox | 86 | Single | Sagittal | 14, 17, 21 & 24 GW |
| | 2 | 336 | SquareBBox | 86 | Single | Sagittal | 14, 17, 21 & 24 GW |
| | 3 | 336 | SquareBBox | 86 | Multi | Sagittal | 14, 17, 21 & 24 GW |
| | 4 | 336 | SquareBBox | 60 | Single | Sagittal | 24 GW |
| | 5 | 336 | ExactBBox | 60 | Single | Sagittal | 24 GW |
| | 6 | 336 | ExactBBox with mask | 60 | Single | Sagittal | 24 GW |
| | 7 | 224 | SquareBBox | 57 | Single | Coronal | 22 GW |
| High Res. Tiles | 8 | 224 | SquareBBox | 379 | Single | Sagittal | 14, 17, 21 & 24 GW |
| | 9 | 224 | SquareBBox | 300 | Single | Coronal | 22 GW |

TABLE 2
COMPARISON OF REGION CLASSIFICATION PERFORMANCE BETWEEN BASE MODELS (ZERO SHOT) AND CYTOCLIP MODELS. THE BASE MODELS FOR CYTOCLIP MODELS ARE SHOWN INSIDE THE PARENTHESIS.

| Image Type | Model Type | Dataset # | Model Name | Training Labels | Label Type | Weighted F1 | Macro F1 | Micro Accuracy |
|---|---|---|---|---|---|---|---|---|
| Complete Region | Base | 2 | CLIP | 86 | Single | 0.0005 | 0.0003 | 0.0041 |
| | | 1 | BiomedCLIP | 86 | Single | 0.004 | 0.002 | 0.028 |
| | Our Model | 2 | CytoCLIP (CLIP) | 86 | Single | 0.839 | **0.652** | 0.834 |
| | | 1 | CytoCLIP (CLIP) | 86 | Single | 0.855 | 0.610 | 0.847 |
| | | 1 | CytoCLIP (BiomedCLIP) | 86 | Single | 0.875 | **0.644** | 0.873 |
| | | 3 | CytoCLIP (CLIP) | 86 | Multi | **0.932** | 0.323 | **0.913** |
| High Res. Tiles | Our Model | 8 | CytoCLIP (BiomedCLIP) | 379 | Single | **0.912** | 0.568 | **0.909** |

## IV. RESULTS AND DISCUSSION

The details of the datasets used to evaluate the performance of the models are shown in Table 1. Wherever applicable, the corresponding dataset number has been cited along with the evaluation result.

### *A. Zero-shot performance of the base models*

The zero-shot classification accuracy of the base models on our brain dataset is significantly low, close to zero, as shown in Table 2. Both OpenAI CLIP and BiomedCLIP models were trained on publicly available image-text pairs. One possible reason for their underperformance is that they haven't encountered similar image-text pairs, specifically NISSL-stained images of brain regions, during training.

The base BiomedCLIP model demonstrates slightly better zero-shot classification accuracy than the OpenAI models. However, we observed that it can correctly classify the "*Hippocampal formation*" regions. For "*Hippocampal formation*", the classification accuracy is 71/88, i.e., 80.68%. The BiomedCLIP has been trained on the PMC-15M dataset (Zhang et al., 2025), which likely included NISSL-stained image-text pairs during its pre-training phase. The precision / recall values for all the classification tasks are provided in the *Supplementary material*.

TABLE 3
REGION CLASSIFICATION PERFORMANCE ACROSS DIFFERENT CROPPING METHODS

| Dataset # | Image Type | Weighted F1 | Macro F1 | Micro Accuracy |
|---|---|---|---|---|
| 5 | ExactBBox | 0.796 | 0.703 | 0.775 |
| 6 | ExactBBox with mask | 0.677 | 0.529 | 0.648 |
| 4 | SquareBBox | **0.799** | **0.732** | **0.789** |

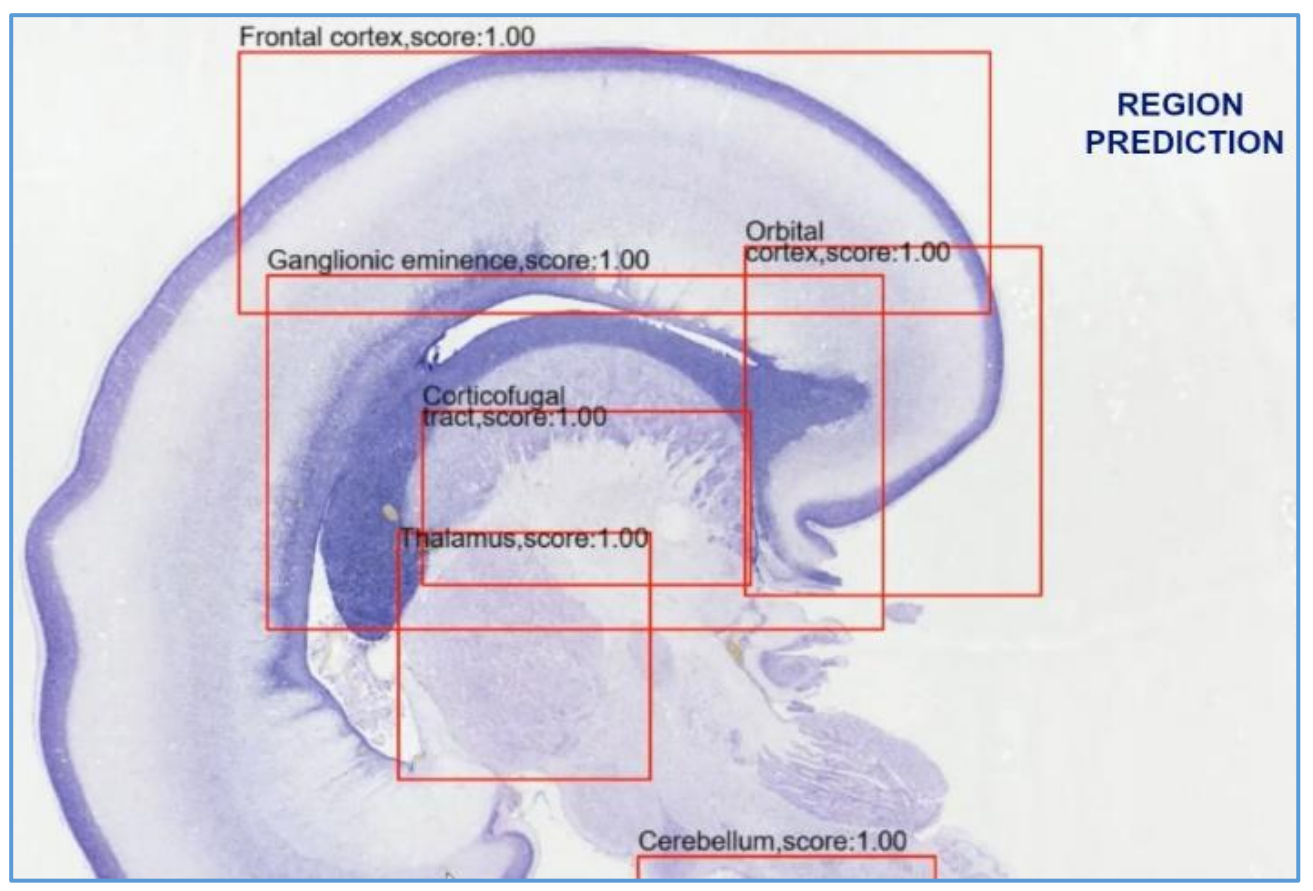

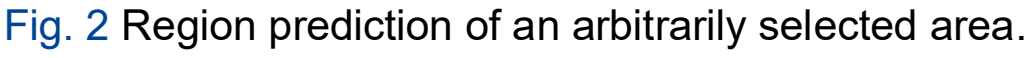

Fig. 2 Region prediction of an arbitrarily selected area.

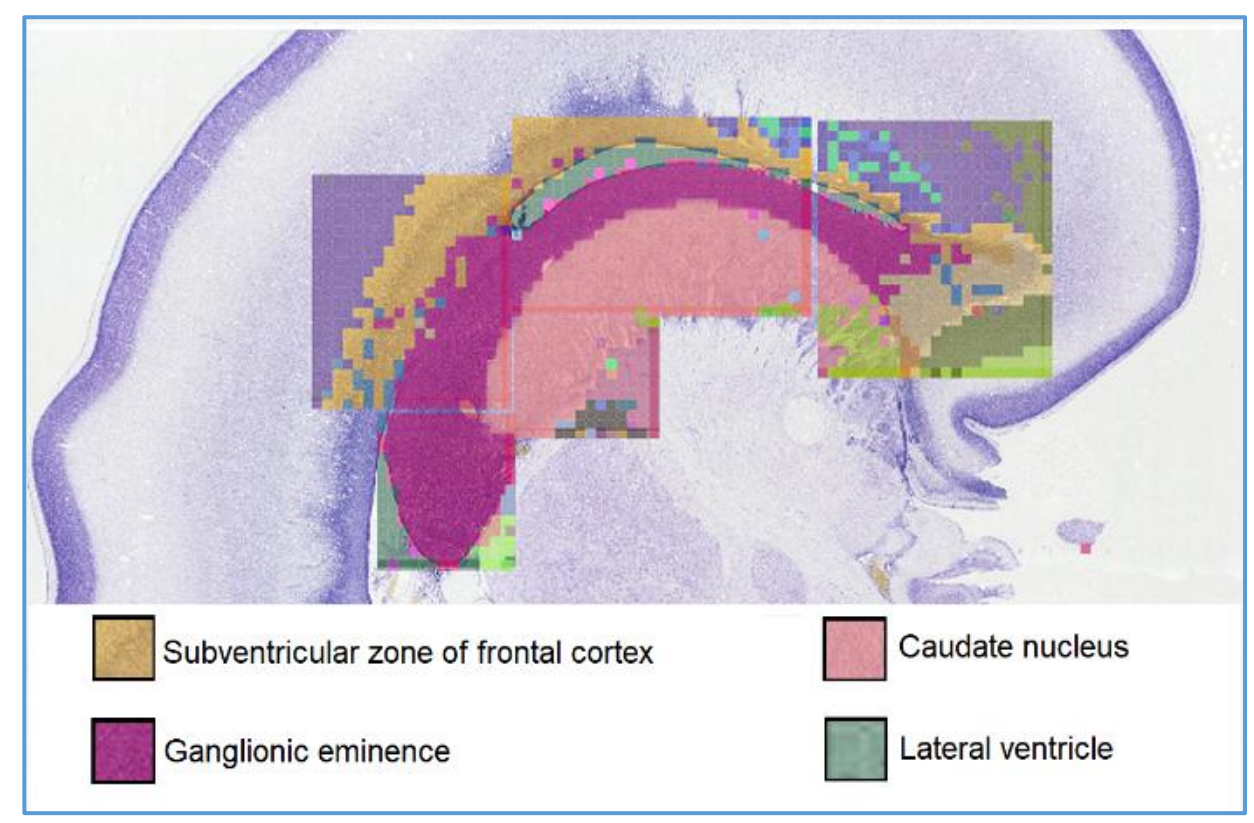

Fig. 3 Coarse-grained segmentation of the whole slide image (only four region names are shown).

## B. Performance of CytoCLIP models for the whole region images

As detailed in the methods section, we generated three distinct types of images, i.e., **ExactBBox, ExactBBox with mask**, and **SquareBBox**. First, we evaluated which type of image yields better classification accuracy using a small portion of the dataset, having 1184 images with 60 different region types. We used OpenAI's clip-vit-large-patch14-336 model for this purpose. Table 3 shows the validation scores of the trained model for the three image types. As we can see, the classification accuracy is higher for the **SquareBBox** types of images. We have used these images in all our subsequent experiments, mentioned below.

Table 2 shows the brain region classification scores for CytoCLIP models trained on image data from the histological brain sections generated using Sagittal cross-sectioning. The models achieved a weighted F1 score of more than 0.83 and a macro F1 score of more than 0.61 for whole-region images. The results indicate that the models successfully predict the names of the majority of the brain regions by learning their structural characteristics; however, they struggle to generalize this learning across all regions. As shown in Fig. 2, the models are employed in a downstream application to predict the name of an arbitrarily selected area within a complete histological section, along with a prediction score.

The difference between weighted F1 and Macro F1 indicates that regions with large support in the dataset, such as *cortical regions* and *ganglionic eminence*, have higher F1 scores. Also, the lower macro-averaged F1 score indicates that some regions have very low F1 scores. For the classification results in Table 1, the corresponding confusion matrices, either full or filtered, are provided in the *Supplementary material (figures S1-S7)*. The filtered confusion matrix has been derived from the complete confusion matrix by using the top 20 region labels with the highest error counts, i.e., the most confused labels. For whole-region images, the matrices (*Supplementary figures S3-S5*) show several regions with significantly lower percentages of true positive values or, in some cases, none at all. This observation clarifies the reasons behind the low macro-averaged F1 scores.

Our analysis found that a significant number of classification errors occur in brain regions that are adjacent to one another. Such neighboring regions often appear in the same image, and the models misclassify one region as another, e.g., Hippocampus and Subiculum or Hippocampus and Corpus callosum. As described in the methodology section, we created an additional dataset with a multi-region label, which includes the names of neighboring regions in the label to address this issue. We then checked whether the primary label appeared among the predicted labels for these images. The weighted F1 score for this revised dataset is the highest, as shown in Table 2, but the macro F1 score is significantly low.

### C. Performance of CytoCLIP models for the WSI tiles

The high-resolution image tile dataset contains a total of 379 distinct labels, representing different brain regions. This large number of labels makes the classification task significantly complex. Despite this complexity, the CytoCLIP model achieved a weighted F1 score of 0.91 and a macro F1 score of 0.57, as shown in Table 2. This score indicates that although the model achieves high overall accuracy on high-resolution image tiles, it does not uniformly learn the detailed cytoarchitecture across the dataset. The difference between the weighted F1 score and the macro F1 score is more pronounced for high-resolution tiles than for whole-region images. Because the frequently occurring regions in the brain sections, such as the *cortical areas* and the *ganglionic eminence*, also occupy large section areas. Hence, the number of tiles for such regions is much higher in the dataset compared to others. The high-resolution tile-based classification task enables a coarse-grained segmentation of the whole slide image by assigning the predicted region label to all pixels in that tile, as shown in Fig. 3.

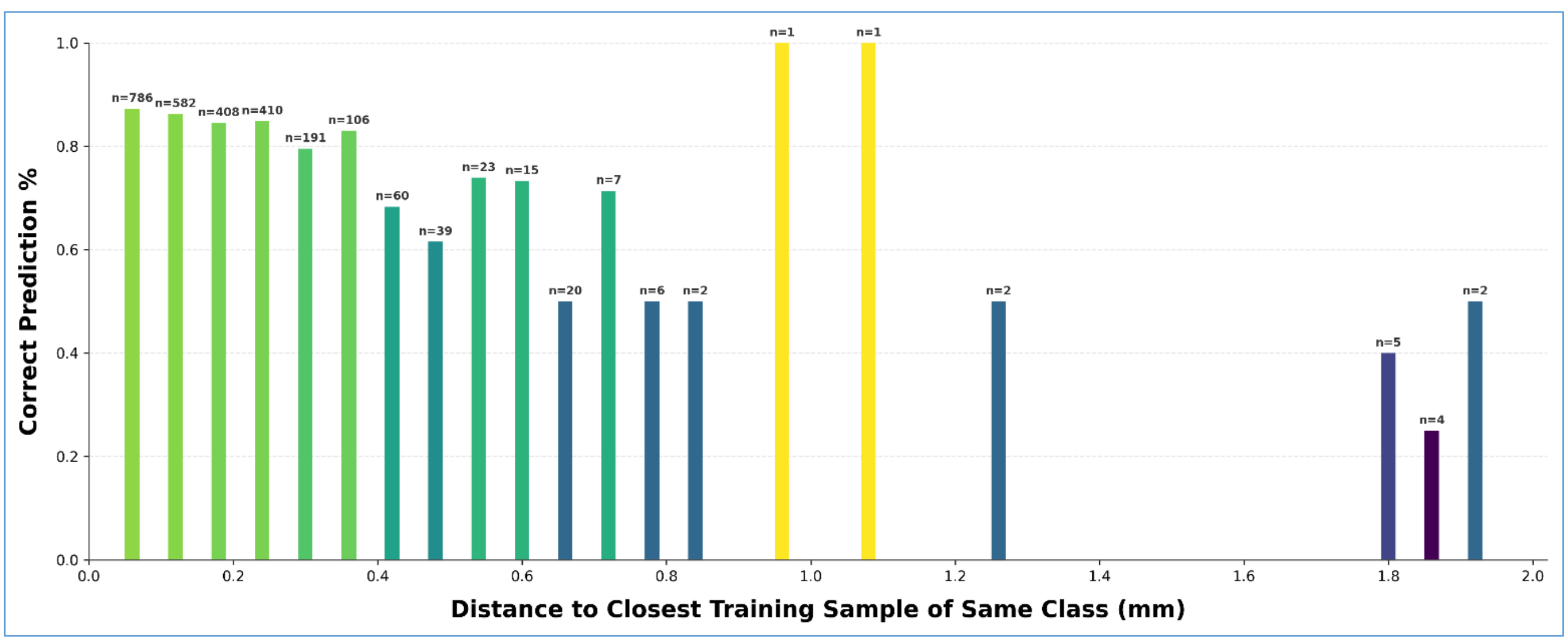

Fig. 4 Percentage of correct predictions across distance to the nearest training sample.

### D. Classification performance across spatial distances

We examined how the spatial distance between a test sample and the nearest training sample of the same class affects classification performance. The images from the test split of the whole-region dataset #2 were used for this analysis. For a test image, we identified the closest brain section that contained a training sample from the same class. In the DHARANI dataset, each brain section has a thickness of 20 µm. The percentages of correct predictions, based on the distance to the closest training sample, are shown in Fig. 4. The numbers displayed above the bars for each distance bin indicate the number of test samples that have their nearest training sample at that specific distance. It is evident from these figures that the percentage of correct predictions decreases as the distance increases, except for a few exceptions. These results indicate that the characteristics of the region images change as the distance between the brain sections increases.

TABLE 4

REGION CLASSIFICATION PERFORMANCE: ACROSS BRAIN SAMPLE AGES AND SECTIONING PLANES.

| Image Type | Dataset # | Training Data | Test Data | Weighted F1 | Macro F1 | Micro Accuracy |
|---|---|---|---|---|---|---|
| Complete Region | 2 | 24 GW (Sagittal) | 21 GW (Sagittal) | 0.365 | 0.204 | 0.342 |
| | 1 & 7 | 14, 17, 21 & 24 GW (Sagittal) | 22 GW (Coronal) | **0.385** | **0.241** | **0.376** |
| High Res. Tiles | 8 & 9 | 14, 17, 21 & 24 GW (Sagittal) | 22 GW (Coronal) | 0.257 | 0.060 | 0.249 |

TABLE 5

RETRIEVAL PERFORMANCE OF CYTOCLIP AND OTHER MODELS IN DIFFERENT MODALITIES. THE BASE MODELS FOR CYTOCLIP MODELS ARE SHOWN INSIDE THE PARENTHESIS.

| Image Type | Model Name | Image to Image | | | Text to Image | | | Image to Text | | |
|---|---|---|---|---|---|---|---|---|---|---|
| | | **Rec. @1** | **Rec. @5** | **Rec. @10** | **Rec. @1** | **Rec. @5** | **Rec. @10** | **Rec. @1** | **Rec. @5** | **Rec. @10** |
| Complete Region | CLIP (Base) | 0.019 | 0.061 | 0.089 | 0.001 | 0.007 | 0.012 | 0.001 | 0.004 | 0.009 |
| | BiomedCLIP (Base) | 0.021 | 0.071 | 0.108 | 0.001 | 0.007 | 0.016 | 0.001 | 0.005 | 0.010 |
| | UNI | 0.024 | 0.081 | 0.119 | -- | -- | -- | -- | -- | -- |
| | CONCH | 0.021 | 0.067 | 0.103 | 0.002 | 0.005 | 0.008 | 0.001 | 0.006 | 0.014 |
| | CytoCLIP (CLIP) | **0.045** | **0.197** | **0.34** | **0.055** | **0.237** | **0.400** | **0.046** | **0.214** | **0.375** |
| | CytoCLIP (BiomedCLIP) | **0.048** | **0.204** | **0.36** | **0.057** | **0.246** | **0.422** | **0.048** | **0.224** | **0.393** |
| High Res. Tiles | CLIP (Base) | 0.019 | 0.057 | 0.089 | 0 | 0.001 | 0.001 | 0 | 0 | 0.001 |
| | BiomedCLIP (Base) | 0.017 | 0.059 | 0.094 | 0.001 | 0.004 | 0.009 | 0 | 0 | 0 |
| | UNI | 0.028 | 0.096 | 0.146 | -- | -- | -- | -- | -- | -- |
| | CONCH | 0.021 | 0.076 | 0.119 | 0 | 0 | 0.002 | 0 | 0.003 | 0.005 |
| | CytoCLIP (BiomedCLIP) | **0.044** | **0.17** | **0.275** | **0.053** | **0.191** | **0.302** | **0.056** | **0.189** | **0.296** |
| BRACS | CLIP (Base) | 0.016 | 0.074 | 0.131 | -- | -- | -- | -- | -- | -- |
| | BiomedCLIP (Base) | 0.019 | 0.075 | 0.138 | -- | -- | -- | -- | -- | -- |
| | UNI | 0.022 | 0.087 | 0.149 | -- | -- | -- | -- | -- | -- |
| | CONCH | **0.023** | **0.096** | **0.170** | -- | -- | -- | -- | -- | -- |
| | CytoCLIP (CLIP) | 0.013 | 0.050 | 0.090 | -- | -- | -- | -- | -- | -- |
| | CytoCLIP (BiomedCLIP) | 0.014 | 0.065 | 0.116 | -- | -- | -- | -- | -- | -- |

### *E. Performance of CytoCLIP models across the different brain sample and sectioning planes*

We investigated whether a model trained on histological section images from one fetal brain can also understand the section images from another. In the DHARANI dataset, the fetal brain sample S5 has an age of 24 gestational weeks (GW), while sample S3 is 21 GW old. Both samples were sectioned in the sagittal plane. We trained a CLIP model using sample S5 and then used it to predict regions in sample S3. Table 4 shows the model's performance; the model achieved a weighted F1 score of 0.37 and a macro F1 score of 0.20. This result suggests that the model couldn't correctly classify many regions, indicating a significant difference in the characteristics of the region images between the two samples.

We also explored whether a model trained on sagittal-sectioned images can understand coronal-sectioned images. To do this, we used the model trained on all the sagittal-sectioned region images from the four brain samples to predict the regions of the brain sectioned in the coronal plane. Table 4 shows the model's performance for the whole region images and WSI tiles. In this scenario, we also observed that the model couldn't accurately predict the names of many regions. It achieved a weighted F1 score of 0.38 and a macro F1 score of 0.24 for the whole region images. For the WSI tiles, the F1 scores are even lower, i.e., 0.26 and 0.06, indicating a significant difference in the regions' visible cellular pattern when sectioned in different planes. Additionally, the macro F1 score, which is much lower than the weighted F1 score, indicates that only a few frequently occurring regions are being accurately identified.

### *F. Retrieval performance in different modalities*

We evaluated the cross-modal retrieval performance of CytoCLIP models and compared it against the SOTA models in the histopathology domain, CONCH (Lu et al., 2024) & UNI (R. J. Chen et al., 2024). As in CONCH (Lu et al., 2024), we used the Recall@K metric, with K = 1, 5, and 10, to evaluate performance for three retrieval tasks: image-to-image, text-to-image, and image-to-text, as depicted in Fig. 1(b) and 1(c). Recall@K measures the fraction of all relevant items that are present in the top K items. We used two datasets for the cross-modal retrieval task. The first dataset was generated by randomly sampling a subset of the test dataset discussed in the methodology section. It contains 925 whole-region images representing 75 distinct regions and 1237 WSI tile images from 373 different brain regions. The second dataset was generated from the testing subset of the BRACS

ROI dataset (Brancati et al., 2022). The BRACS dataset contains hematoxylin and eosin (H&E) stained histopathological images of 6 breast cancer lesion subtypes and normal tissue. We randomly selected 193 test ROI images representing all the 7 labels.

As shown in Table 5, CytoCLIP outperforms the base models as well as the SOTA models in all three retrieval tasks for the first dataset. For whole-region images, CytoCLIP, based on BiomedCLIP, performs best among all the models. The performance gap between CytoCLIP and other tested models is significant for cross-modal retrieval tasks as compared to image-to-image retrieval tasks. This result suggests that while other tested models can create embeddings that are closer for similar NISSL-stained brain histological images, the embeddings for image-text pairs in the shared embedding space do not have a closer proximity to each other. Also, among all the other tested models, UNI (R. J. Chen et al., 2024) has the highest image-to-image retrieval performance.

For the BRACS dataset, only image-to-image retrieval was evaluated, as neither CytoCLIP nor CONCH was trained on the BRACS labels. CONCH outperforms other models for this task. The performance of CytoCLIP models is lower than that of the base models, indicating they are not generalizing to H&E-stained histopathological images.

## V. Conclusion

In this work, our research focused on learning the cytoarchitectural characteristics of different brain regions in histological sections of fetal brains using Contrastive Language-Image Pre-Training (CLIP). We prepared two datasets: images of complete regions and high-resolution image tiles from brain histological section images. We train a set of models, based on the CLIP framework, using our dataset. Our experiments indicate that the models can effectively identify different brain regions, suggesting they learn some distinct cytoarchitectural characteristics. However, the variation in performance across all regions indicates that identifying more subtle cytoarchitectural differences remains a challenge for the models. Our experiments also show that the models outperform other state-of-the-art models on various retrieval tasks. We believe that the ability to understand low-resolution images of complete regions will help to interpret other low-resolution NISSL-stained brain histological images available in the public domain, especially those used in the neuroscience literature. This capability will help integrate multimodal data. Whereas learning from high-resolution image tiles can enable automatic delineation of brain regions.

However, based on our analysis of the results presented above, several areas require more research for further improvement. We observed that the macro F1 classification score is lower than the weighted F1 score, indicating a need for further improvement. Additionally, models trained on images generated with one sectioning plane do not generalize well to images generated with a different sectioning plane. Also, the models show limited generalization ability across brain samples of various ages and across other histopathological images. More research is required to improve the generalization ability of models across sectioning planes, brain sample ages, and various histopathological images. Lastly, the results of coarse-grained segmentation need to be analyzed and reported in greater detail using quantitative metrics.

## References

Amunts, K., & Zilles, K. (2015). Architectonic mapping of the human brain beyond brodmann. *Neuron*, *88*(6), 1086–1107. https://doi.org/10.1016/j.neuron.2015.12.001

Brancati, N., Anniciello, A. M., Pati, P., Riccio, D., Scognamiglio, G., Jaume, G., De Pietro, G., Di Bonito, M., Foncubierta, A., Botti, G., Gabrani, M., Feroce, F., & Frucci, M. (2022). BRACS: A dataset for BReAst Carcinoma Subtyping in H&E histology images. Database: The Journal of Biological Databases and Curation, 2022. https://doi.org/10.1093/database/baac093

Chen, R. J., Chen, C., Li, Y., Chen, T. Y., Trister, A. D., Krishnan, R. G., & Mahmood, F. (2022). Scaling vision transformers to gigapixel images via hierarchical self-supervised learning. *2022 IEEE/CVF Conference on Computer Vision and Pattern Recognition (CVPR)*.

Chen, R. J., Ding, T., Lu, M. Y., Williamson, D. F. K., Jaume, G., Song, A. H., Chen, B., Zhang, A., Shao, D., Shaban, M., Williams, M., Oldenburg, L., Weishaupt, L. L., Wang, J. J., Vaidya, A., Le, L. P., Gerber, G., Sahai, S., Williams, W., & Mahmood, F. (2024). Towards a general-purpose foundation model for computational pathology. *Nature Medicine*, *30*(3), 850–862. https://doi.org/10.1038/s41591-024-02857-3

Chen, T., Kornblith, S., Norouzi, M., & Hinton, G. (2020). A simple framework for contrastive learning of visual representations. *Proceedings of the 37th International Conference on Machine Learning*, 1597–1607.

Ciga, O., Xu, T., & Martel, A. L. (2022). Self supervised contrastive learning for digital histopathology. *Machine Learning With Applications*, *7*(100198), 100198. https://doi.org/10.1016/j.mlwa.2021.100198

Das, A., Chaudhuri, M., Bhat, K., Ram, K., Bota, M., & Sivaprakasam, M. (2025). PosDiffAE: Position-aware diffusion auto-encoder for high-resolution brain tissue classification incorporating artifact restoration. *IEEE Journal of Biomedical and Health Informatics*, *PP*. https://doi.org/10.1109/JBHI.2025.3567708

Ding, S.-L., Royall, J. J., Lesnar, P., Facer, B. A. C., Smith, K. A., Wei, Y., Brouner, K., Dalley, R. A., Dee, N., Dolbeare, T. A., Ebbert, A., Glass, I. A., Keller, N. H., Lee, F., Lemon, T. A., Nyhus, J., Pendergraft, J., Reid, R., Sarreal, M., … Lein, E. S. (2022). Cellular resolution anatomical and molecular atlases for prenatal human brains. *The Journal of Comparative Neurology*, *530*(1), 6–503. https://doi.org/10.1002/cne.25243

Ding, S.-L., Royall, J. J., Sunkin, S. M., Ng, L., Facer, B. A. C., Lesnar, P., Guillozet-Bongaarts, A., McMurray, B., Szafer, A., Dolbeare, T. A., Stevens, A., Tirrell, L., Benner, T., Caldejon, S., Dalley, R. A., Dee, N., Lau, C., Nyhus, J., Reding, M., … Lein, E. S. (2016). Comprehensive cellular-resolution atlas of the adult human brain. *The Journal of Comparative Neurology*, *524*(16), 3127–3481. https://doi.org/10.1002/cne.24080

Eslami, S., Meinel, C., & de Melo, G. (2023). PubMedCLIP: How much does CLIP benefit visual question answering in the medical domain? *Findings of the Association for Computational Linguistics: EACL 2023*, 1181–1193.

Fashi, P. A., Hemati, S., Babaie, M., Gonzalez, R., & Tizhoosh, H. R. (2022). A self-supervised contrastive learning approach for whole slide image representation in digital pathology. *Journal of Pathology Informatics*, *13*(100133), 100133. https://doi.org/10.1016/j.jpi.2022.100133

Hu, W., Li, X., Li, C., Li, R., Jiang, T., Sun, H., Huang, X., Grzegorzek, M., & Li, X. (2023). A state-of-the-art survey of artificial neural networks for Whole-slide Image analysis: From popular Convolutional Neural Networks to potential visual transformers. *Computers in Biology and Medicine*, *161*(107034), 107034. https://doi.org/10.1016/j.compbiomed.2023.107034

Huang, Z., Bianchi, F., Yuksekgonul, M., Montine, T. J., & Zou, J. (2023). A visual-language foundation model for pathology image analysis using medical Twitter. *Nature Medicine*, *29*(9), 2307–2316. https://doi.org/10.1038/s41591-023-02504-3

Ilharco, G., Wortsman, M., Wightman, R., Gordon, C., Carlini, N., Taori, R., Dave, A., Shankar, V., Namkoong, H., Miller, J., Hajishirzi, H., Farhadi, A., & Schmidt, L. (2025). *OpenCLIP*. Zenodo.

Kakkar, M., Shanbhag, D., Aladahalli, C., & M, G. R. (2024). Language augmentation in CLIP for improved anatomy detection on multi-modal medical images. *Annual International Conference of the IEEE Engineering in Medicine and Biology Society. IEEE Engineering in Medicine and Biology Society. Annual International Conference*, *2024*, 1–4. https://doi.org/10.1109/EMBC53108.2024.10781689

Komura, D., & Ishikawa, S. (2018). Machine learning methods for histopathological image analysis. *Computational and Structural Biotechnology Journal*, *16*, 34–42. https://doi.org/10.1016/j.csbj.2018.01.001

Kosaraju, S., Park, J., Lee, H., Yang, J. W., & Kang, M. (2022). Deep learning-based framework for slide-based histopathological image analysis. *Scientific Reports*, *12*(1), 19075. https://doi.org/10.1038/s41598-022-23166-0

Loshchilov, I., & Hutter, F. (2019). Decoupled Weight Decay Regularization. *7th International Conference on Learning Representations, ICLR 2019, New Orleans, LA, USA, May 6-9, 2019*. https://openreview.net/forum?id=Bkg6RiCqY7

Lu, M. Y., Chen, B., Williamson, D. F. K., Chen, R. J., Liang, I., Ding, T., Jaume, G., Odintsov, I., Le, L. P., Gerber, G., Parwani, A. V., Zhang, A., & Mahmood, F. (2024). A visual-language foundation model for computational pathology. *Nature Medicine*, *30*(3), 863–874. https://doi.org/10.1038/s41591-024-02856-4

Lu, M. Y., Chen, B., Zhang, A., Williamson, D. F. K., Chen, R. J., Ding, T., Le, L. P., Chuang, Y.-S., & Mahmood, F. (2023). Visual language pretrained multiple instance zero-shot transfer for histopathology images. *2023 IEEE/CVF Conference on Computer Vision and Pattern Recognition (CVPR)*.

Madabhushi, A., & Lee, G. (2016). Image analysis and machine learning in digital pathology: Challenges and opportunities. *Medical Image Analysis*, *33*, 170–175. https://doi.org/10.1016/j.media.2016.06.037

Patel, T., El-Sayed, H., & Sarker, M. K. (2024). Evaluating Vision-Language Models for hematology image Classification: Performance Analysis of CLIP and its Biomedical AI Variants. *2024 36th Conference of Open Innovations Association (FRUCT)*, 578–584.

Radford, A., Kim, J. W., Hallacy, C., Ramesh, A., Goh, G., Agarwal, S., Sastry, G., Askell, A., Mishkin, P., Clark, J., Krueger, G., & Sutskever, I. (2021). Learning Transferable Visual Models From Natural Language Supervision. In M. Meila & T. Zhang (Eds.), *Proceedings of the 38th International Conference on Machine Learning* (Vol. 139, pp. 8748–8763). PMLR.

Rahaman, M. M., Millar, E. K. A., & Meijering, E. (2025). Generalized deep learning for histopathology image classification using supervised contrastive learning. *Journal of Advanced Research*, *75*, 389–404. https://doi.org/10.1016/j.jare.2024.11.013

Rottschy, C., Eickhoff, S. B., Schleicher, A., Mohlberg, H., Kujovic, M., Zilles, K., & Amunts, K. (2007). Ventral visual cortex in humans: cytoarchitectonic mapping of two extrastriate areas. *Human Brain Mapping*, *28*(10), 1045–1059. https://doi.org/10.1002/hbm.20348

Schiffer, C., Amunts, K., Harmeling, S., & Dickscheid, T. (2021). Contrastive representation learning for whole brain cytoarchitectonic mapping in histological human brain sections. *2021 IEEE 18th International Symposium on Biomedical Imaging (ISBI)*.

Schiffer, C., Spitzer, H., Kiwitz, K., Unger, N., Wagstyl, K., Evans, A. C., Harmeling, S., Amunts, K., & Dickscheid, T. (2021). Convolutional neural networks for cytoarchitectonic brain mapping at large scale. *NeuroImage*, *240*(118327), 118327. https://doi.org/10.1016/j.neuroimage.2021.118327

SGBC-IITM. (2025). *DHARANI Dataset*. Sudha Gopalakrishnan Brain Centre IIT Madras. https://brainportal.humanbrain.in/publicview/index.html

Spitzer, H., Amunts, K., Harmeling, S., & Dickscheid, T. (2017). Parcellation of visual cortex on high-resolution histological brain sections using convolutional neural networks. *2017 IEEE 14th International Symposium on Biomedical Imaging (ISBI 2017)*.

Spitzer, H., Kiwitz, K., Amunts, K., Harmeling, S., & Dickscheid, T. (2018). Improving cytoarchitectonic segmentation of human brain areas with self-supervised Siamese networks. In *Medical Image Computing and Computer Assisted Intervention – MICCAI 2018* (pp. 663–671). Springer International Publishing.

Tan, J. W., & Jeong, W.-K. (2023). Histopathology image classification using deep manifold contrastive learning. In *Lecture Notes in Computer Science* (pp. 683–692). Springer Nature Switzerland.

Verma, R., Bota, M., Ram, K., Jayakumar, J., Folkerth, R., Pandurangan, K., Ramesh, J. J., Majumder, M., Raveendran, R., Nanda, R., K, S., S, A. D., Karthik, S., Kumarasami, R., S, S., Lata, S., Kumar, E. H., Rangasami, R., Srinivasan, C., … Sivaprakasam, M. (2025). DHARANI: A 3D developing human-brain atlas resource to advance neuroscience internationally integrated multimodal imaging and high-resolution histology of the second trimester. *The Journal of Comparative Neurology*, *533*(2), e70006. https://doi.org/10.1002/cne.70006

Wang, Z., Wu, Z., Agarwal, D., & Sun, J. (2022). MedCLIP: Contrastive learning from unpaired medical images and text. *Proceedings of the Conference on Empirical Methods in Natural Language Processing. Conference on Empirical Methods in Natural Language Processing*, *2022*, 3876–3887. https://doi.org/10.18653/v1/2022.emnlp-main.256

Wei, X., Kurtz, C., & Cloppet, F. (2025). Enhancing vision–language contrastive representation learning using domain knowledge. *Computer Vision and Image Understanding*, *259*(104403), 104403. https://doi.org/10.1016/j.cviu.2025.104403

Weiner, K. S., Barnett, M. A., Lorenz, S., Caspers, J., Stigliani, A., Amunts, K., Zilles, K., Fischl, B., & Grill-Spector, K. (2017). The cytoarchitecture of domain-specific regions in human High-level visual cortex. *Cerebral Cortex (New York, N.Y.: 1991)*, *27*(1), 146–161. https://doi.org/10.1093/cercor/bhw361

Yang, P., Hong, Z., Yin, X., Zhu, C., & Jiang, R. (2021). Self-supervised visual representation learning for histopathological images. In *Lecture Notes in Computer Science* (pp. 47–57). Springer International Publishing.

Zhang, S., Xu, Y., Usuyama, N., Xu, H., Bagga, J., Tinn, R., Preston, S., Rao, R., Wei, M., Valluri, N., Wong, C., Tupini, A., Wang, Y., Mazzola, M., Shukla, S., Liden, L., Gao, J., Crabtree, A., Piening, B., … Poon, H. (2025). A multimodal biomedical foundation model trained from fifteen million image–text pairs. *NEJM AI*, *2*(1). https://doi.org/10.1056/aioa2400640

Zhao, F., Wang, Z., Du, H., He, X., & Cao, X. (2023). Self-supervised triplet contrastive learning for classifying endometrial histopathological images. *IEEE Journal of Biomedical and Health Informatics*, *27*(12), 5970–5981. https://doi.org/10.1109/JBHI.2023.3314663

Zhao, Z., Liu, Y., Wu, H., Wang, M., Li, Y., Wang, S., Teng, L., Liu, D., Cui, Z., Wang, Q., & Shen, D. (2025). CLIP in medical imaging: A survey. *Medical Image Analysis*, *102*(103551), 103551. https://doi.org/10.1016/j.media.2025.103551

Zilles, K., & Amunts, K. (2010). Centenary of Brodmann's map--conception and fate. *Nature Reviews. Neuroscience*, *11*(2), 139–145. https://doi.org/10.1038/nrn2776